\documentclass[conference]{IEEEtran}
\IEEEoverridecommandlockouts
\usepackage{cite}
\usepackage{stfloats}
\usepackage[numbers,sort&compress]{natbib}
\usepackage[most,listings]{tcolorbox}
\usepackage{xcolor}  % 如果需要自定义颜色
\usepackage[utf8]{inputenc}
\usepackage{amsmath,amssymb}
\usepackage{graphicx}
\usepackage{bm}
\usepackage{enumerate}
\usepackage{comment}
\usepackage{xcolor}
\usepackage{url}
\usepackage{color,soul}
\usepackage{subcaption}
\usepackage{float}
\usepackage{booktabs} % For prettier tables

\definecolor{light-gray}{gray}{0.8}

\usepackage{amsmath,amssymb,amsfonts}
\usepackage{algorithmic}
\usepackage{graphicx}
\usepackage{textcomp}
\usepackage{xcolor}
\usepackage{subcaption}

\def\BibTeX{{\rm B\kern-.05em{\sc i\kern-.025em b}\kern-.08em
    T\kern-.1667em\lower.7ex\hbox{E}\kern-.125emX}}

\makeatletter
\newcommand{\linebreakand}{%
  \end{@IEEEauthorhalign}
  \hfill\mbox{}\par
  \mbox{}\hfill\begin{@IEEEauthorhalign}
}
\makeatother

\tcbset{
  myprompt/.style={
    colback=gray!10,
    colframe=gray!50,
    fonttitle=\bfseries,
    title=Prompt without CoT,
    boxrule=0.4pt,
    arc=3pt,
    outer arc=3pt,
    top=1mm,
    bottom=1mm,
    left=1mm,
    right=1mm,
  },
  myquestion/.style={
    colback=blue!5,
    colframe=blue!40,
    fonttitle=\bfseries,
    title=Prompt with CoT,
    boxrule=0.4pt,
    arc=3pt,
    outer arc=3pt,
    top=1mm,
    bottom=1mm,
    left=1mm,
    right=1mm,
  },
  myanswer/.style={
    colback=green!5,
    colframe=green!40,
    fonttitle=\bfseries,
    title=LLM Answer,
    boxrule=0.4pt,
    arc=3pt,
    outer arc=3pt,
    top=1mm,
    bottom=1mm,
    left=1mm,
    right=1mm,
  }
}

\begin{document}

\title{Instruction Tuning and CoT Prompting for Contextual Medical QA with LLMs}

\author{

\small % Set font size to 10pt

\begin{tabular}[t]{c@{\extracolsep{8em}}c} 

1\textsuperscript{st} Chenqian Le\textsuperscript{*} & 2\textsuperscript{nd} Ziheng Gong \\
\textit{New York University, New York, USA} & \textit{New York University, New York, USA} \\
\textsuperscript{*}Corresponding author: cl6707@nyu.edu \\

\\

3\textsuperscript{rd} Chihang Wang & 4\textsuperscript{th} Haowei Ni \\
\textit{New York University, New York, USA} & \textit{Columbia University, New York, USA} \\

\\

5\textsuperscript{th} Panfeng Li & 6\textsuperscript{th}  Xupeng Chen \\
\textit{University of Michigan, Ann Arbor, USA} & \textit{New York University, New York, USA} \\

\end{tabular}
}
\maketitle

\begin{abstract}
Large language models (LLMs) have shown great potential in medical question answering (MedQA), yet adapting them to biomedical reasoning remains challenging due to domain-specific complexity and limited supervision. In this work, we study how prompt design and lightweight fine-tuning affect the performance of open-source LLMs on PubMedQA, a benchmark for multiple-choice biomedical questions. We focus on two widely used prompting strategies—standard instruction prompts and Chain-of-Thought (CoT) prompts—and apply QLoRA for parameter-efficient instruction tuning. Across multiple model families and sizes, our experiments show that CoT prompting alone can improve reasoning in zero-shot settings, while instruction tuning significantly boosts accuracy. However, fine-tuning on CoT prompts does not universally enhance performance and may even degrade it for certain larger models. These findings suggest that reasoning-aware prompts are useful, but their benefits are model- and scale-dependent. Our study offers practical insights into combining prompt engineering with efficient finetuning for medical QA applications.
\end{abstract}

\begin{IEEEkeywords}
Large Language Model, Quantized Low-Rank Adaptation, Instruction Fine Tuning, Chain of Thought, Medical Question Answering
\end{IEEEkeywords}

\section{Introduction}
In the modern healthcare landscape, clinicians, researchers, and patients are inundated with an ever-growing volume of biomedical literature, clinical guidelines, and patient records~\cite{ding2025ai,10628639}. Making timely and accurate decisions often hinges on the ability to extract, interpret, and synthesize complex medical knowledge from this unstructured textual data. Traditional search engines and keyword-based tools struggle to meet the demands of such high-stakes environments, where nuances in language, context, and reasoning can significantly alter clinical outcomes. Moreover, the steep learning curve associated with biomedical jargon makes it challenging for non-experts to engage with medical content effectively.

Recent progress in deep learning has significantly advanced natural language processing (NLP), enabling systems to interpret structured and unstructured data more effectively~\cite{yang2025research,yang2024hades,jin2025adaptive}. While early biomedical NLP systems, such as domain-specific transformers like BioBERT~\cite{lee2020biobert}, brought substantial improvements, they often required large-scale supervised training and lacked flexibility in reasoning and adaptation. In parallel, advances in multimodal deep learning and vision-language alignment—such as image-to-image generation from textual prompts—demonstrate the broader potential of neural networks in context-rich, data-driven reasoning environments. Methods for predictive modeling in time series data~\cite{sui2024ensemble,zhang2025avocado} further illustrate how diverse deep learning strategies are being tailored to complex real-world applications, including those in healthcare~\cite{ji2023prediction,liu2024drugagent,luo2023towards}.

The development of large language models (LLMs) marks a paradigm shift. Trained on massive corpora, LLMs exhibit strong capabilities in language understanding, abstraction, and contextual reasoning~\cite{ji-etal-2024-rag}. These models can perform complex tasks in zero- or few-shot settings and have shown promise in domains such as image processing~\cite{zhang2022covid}. However, the use of LLMs in biomedical contexts raises serious concerns: many are trained on general-domain text, resulting in hallucinations and factual inaccuracies when applied to clinical reasoning. Their lack of transparency further complicates their integration into safety-critical workflows like diagnosis and patient care.

As machine learning techniques evolve, integration with LLMs has fueled next-generation systems that support intelligent reasoning across modalities and domains. Innovations in federated learning for cross-cloud health data privacy~\cite{luo2025crosscloud,yang2025cloud,ji2025cloud} are redefining how models interact with complex data ecosystems. Likewise, work in data augmentation, ensemble forecasting~\cite{sui2024ensemble}, and 3D reconstruction~\cite{ding2025nerf,ding2025neural} underscore the breadth of technical advancements that, while not exclusive to healthcare, offer transferable methodologies applicable to biomedical AI. Within this broader landscape, medical question answering (MedQA) stands out as a high-impact application that demands precise factual grounding, contextual understanding, and explainability—particularly in clinical decision support and patient communication.

Traditional approaches to MedQA relied on smaller, domain-specific models such as BioBERT~\cite{lee2020biobert} and BioMedLM~\cite{bolton2024biomedlm}. These models offer strong lexical understanding but are often limited in their capacity for reasoning or generalization beyond narrow datasets. The emergence of general-purpose LLMs like GPT-4~\cite{achiam2023gpt} and open-source alternatives such as LLaMA3~\cite{grattafiori2024llama} and Qwen2.5~\cite{yang2024qwen2} has changed the paradigm, making it feasible to leverage broad language understanding for specialized biomedical tasks~\cite{xu2024smiles,yue2024biomamba}.

Despite their potential, challenges remain in deploying LLMs for high-stakes tasks such as MedQA. First, most models are not aligned to the medical domain by default, leading to hallucinations or vague responses. Second, biomedical datasets are often limited in scale, making full fine-tuning costly and potentially unstable. Finally, the reasoning paths taken by LLMs remain opaque, hindering explainability and user trust.

To address these challenges, two complementary strategies have emerged. First, instruction fine-tuning aligns LLM behavior with task-specific expectations using lightweight supervised learning. Second, Chain-of-Thought (CoT) prompting\cite{wei2022chain} encourages models to reason explicitly by generating intermediate rationales before answering. Prior work (e.g., Med-PaLM~\cite{singhal2025toward}) has shown the promise of such techniques in improving factuality and safety in medical QA, though primarily in closed or proprietary models.

In this paper, we extend these ideas to open-source LLMs and ask:

\begin{itemize}
\item Can CoT prompting alone improve performance in biomedical QA under zero-shot conditions?
\item How does instruction fine-tuning interact with CoT prompts?
\item Are gains from CoT consistent across different model families and sizes?
\end{itemize}

We explore these questions on the PubMedQA dataset using multiple open-source LLMs, including Llama-3.1-8B, Llama-3.3-70B, and the Qwen2.5 series. Fine-tuning is performed using QLoRA~\cite{dettmers2023qlora} for efficiency. Our findings reveal that while CoT prompts are helpful in some base settings, their benefits after fine-tuning are model-dependent—suggesting the need for more nuanced prompt-model alignment in clinical reasoning tasks.

\section{Data Collection and Preprocessing}
We use the PubMedQA dataset~\cite{jin2019pubmedqa}, a widely used benchmark in biomedical question answering. Each instance consists of a medical research question, a context (abstract from PubMed), and four possible answers: \texttt{A/B/C/D}.

We created two versions of the input prompt. The standard prompt asks the model to choose one of the options directly, while the Chain-of-Thought prompt encourages intermediate reasoning. Examples are provided below.

\begin{tcolorbox}[myprompt, title=Standard Prompt]
You are a medical expert. Based on the context provided, answer the question by selecting one option. Provide your final answer as one of the options A, B, or C. Your final response should only be the letter corresponding to your chosen option.

Context: \{Context\}

Question: Is anorectal endosonography valuable in dyschesia?

Options:
A: yes \quad B: no \quad C: maybe

Answer: A
\end{tcolorbox}

\begin{tcolorbox}[myquestion, title=Chain-of-Thought Prompt]
You are a medical expert. Based on the context provided, answer the question by selecting one option. First, think step by step about the reasoning for your answer. Then provide your final answer as one of the options A, B, or C. Your final response should only be the letter corresponding to your chosen option.

Context: \{Context\}

Question: Is anorectal endosonography valuable in dyschesia?

Options:
A: yes \quad B: no \quad C: maybe

Think: Let me think through this problem step by step to find the correct answer. \{Think\}

Answer: A
\end{tcolorbox}

\section{Method}
\subsection{Framework}
\begin{figure*}[!htbp]
    \centering
    \includegraphics[width=0.85\textwidth]{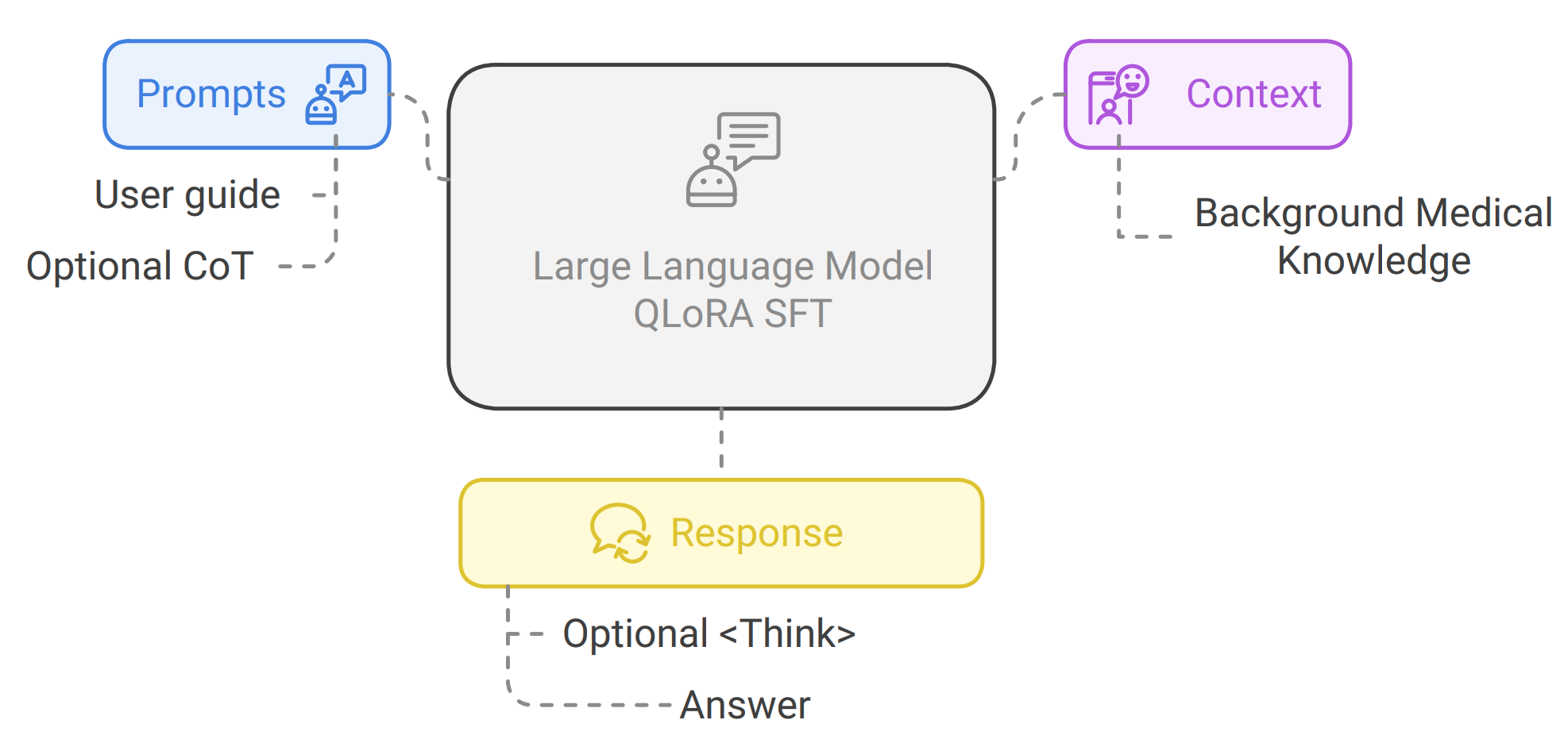}
    \caption{Overview of the framework for medical question answering using LLMs with standard vs. Chain-of-Thought prompting.}
    \label{fig:framework}
\end{figure*}

Our framework, shown in Fig. \ref{fig:framework} supports both standard and CoT prompting. For each model, we perform inference in both base (zero-shot) and instruction fine-tuned (SFT) settings. We train the models using QLoRA for parameter-efficient finetuning and evaluate them using Accuracy and F1 scores on the PubMedQA test set.

\subsection{Instruction Fine Tuning}
Instruction fine-tuning~\cite{zhang2023instruction} is a widely adopted strategy for adapting large language models (LLMs) to domain-specific tasks using relatively small but high-quality supervised datasets. It trains LLMs to follow structured instructions and generate consistent, context-aware responses—particularly beneficial in specialized fields like biomedical QA, where accuracy and reasoning alignment are critical.

Our fine-tuning process consists of three main stages. First, we construct an instruction-following dataset by curating and formatting samples from the PubMedQA training set. Each sample is transformed into a structured prompt-response pair, supporting both standard and Chain-of-Thought (CoT) formats. The CoT variant includes intermediate reasoning steps (prefixed by Think:) as part of the supervision signal.

Second, we apply parameter-efficient fine-tuning using QLoRA~\cite{dettmers2023qlora}, a method that quantizes model weights to 4-bit precision while learning low-rank adaptation matrices. This allows us to efficiently update large models without incurring the prohibitive memory and compute costs of full fine-tuning—enabling training on a single A100 GPU with long input sequences.

Finally, during inference, the fine-tuned model leverages its learned instruction-following capabilities to generate responses that are better aligned with the task requirements. We evaluate the impact of fine-tuning on both accuracy and reasoning quality, especially when combined with CoT-style prompting. This modular fine-tuning pipeline provides a scalable path to domain alignment without sacrificing efficiency.

\subsection{Chain of Thought Prompting}
Prompt design is pivotal in steering large language models (LLMs) toward more structured, interpretable, and context-aware reasoning—an essential feature in high-stakes domains such as biomedicine. Among various strategies, Chain-of-Thought (CoT) prompting has proven effective in guiding models to articulate intermediate reasoning steps before producing a final decision, offering both performance improvements and increased transparency.

In this work, we evaluate the utility of CoT prompting by comparing it with standard instruction-style prompts on multiple-choice biomedical question-answering tasks, where the model must select the most appropriate answer from three or four options: \texttt{A/B/C/D}. Standard prompts instruct the model to directly output a final choice, while CoT prompts add an explicit reasoning phase via a Think: prefix, encouraging the model to reason through the problem step by step before committing to an answer.

While our evaluation focuses on the final selected option for accuracy and F1 scoring, the generated reasoning chains offer additional insight into the model’s internal decision-making process—critical for domains where explainability is valued alongside correctness. Notably, our setup does not include supervised reasoning labels; rather, we leverage the structure of the CoT prompts to induce reasoning behavior in the models.

By systematically comparing both prompt types across various model families and fine-tuning settings, we aim to better understand the benefits and limitations of reasoning-inductive prompting in medical multiple-choice QA.

\subsection{QLoRA}
Quantized Low-Rank Adaptation (QLoRA)~\cite{dettmers2023qlora} represents a state-of-the-art method for efficiently fine-tuning large models by applying quantization to reduce memory and compute usage. We employ 4-bit QLoRA to fine-tune all models, enabling training on a single A100 with long-context sequences.

\subsection{Models and Settings}
We evaluate four open-source LLMs: Llama-3.1-8B, Llama-3.3-70B, Qwen2.5-7B, and Qwen2.5-14B. For each model, we perform inference in two settings:

\begin{itemize}
\item \textbf{Base (Zero-shot)}: Direct inference using the pre-trained checkpoint.
\item \textbf{Instruction Finetuned (SFT)}: Models fine-tuned on formatted PubMedQA data using QLoRA.
\end{itemize}

\subsection{Prompting Strategies}
Two types of prompts are designed:
\begin{itemize}
\item \textbf{Default Prompt}: A concise instruction asking the model to select one of the three options.
\item \textbf{CoT Prompt}: Same as Default but with an explicit \texttt{Think:} section encouraging step-by-step reasoning.
\end{itemize}

During finetuning, we supervise both the reasoning and final answer for CoT. During inference, only the final letter answer is used for evaluation.

\section{Experiments}
\subsection{Training Configuration}

To adapt open-source LLMs to the biomedical domain efficiently, we employed a parameter-efficient fine-tuning strategy using 4-bit Quantized Low-Rank Adaptation (QLoRA)~\cite{dettmers2023qlora}. This approach significantly reduces memory requirements while maintaining model performance, enabling the fine-tuning of large-scale models on a single A100 GPU.

All models were fine-tuned for one epoch on the PubMedQA training set using a consistent configuration to ensure comparability across experiments. Specifically, the optimization process utilized the 8-bit variant of the AdamW optimizer~\cite{loshchilov2017decoupled}, which balances memory efficiency with convergence stability. The initial learning rate was set to $2 \times 10^{-4}$, with a linear scheduler to gradually decay the learning rate over the course of training.

Each training run supported a maximum input sequence length of 25,000 tokens, which is critical for handling the long context passages typical of biomedical abstracts. Due to the high memory demand of long-context fine-tuning, we dynamically adjusted the effective batch size through gradient accumulation. This strategy allowed for stable optimization without exceeding GPU memory limits while still enabling the models to benefit from large-scale sequence input.

While no weight decay or warm-up steps were used in this configuration, we observed that a single epoch of fine-tuning with QLoRA was sufficient to adapt the models effectively to the biomedical QA task. Our pipeline demonstrated that fine-tuning large models with long input contexts is both feasible and effective on limited hardware resources.

\subsection{Metrics}

We report performance using the following evaluation metrics:

\begin{itemize}
\item \textbf{Accuracy:} The proportion of correct predictions overall test instances.
\item \textbf{Weighted F1 Score:} An F1 score weighted by class distribution is particularly important for PubMedQA due to its imbalanced answer types.
\end{itemize}

\begin{figure*}[!htbp]
    \centering
    \includegraphics[width=0.85\textwidth]{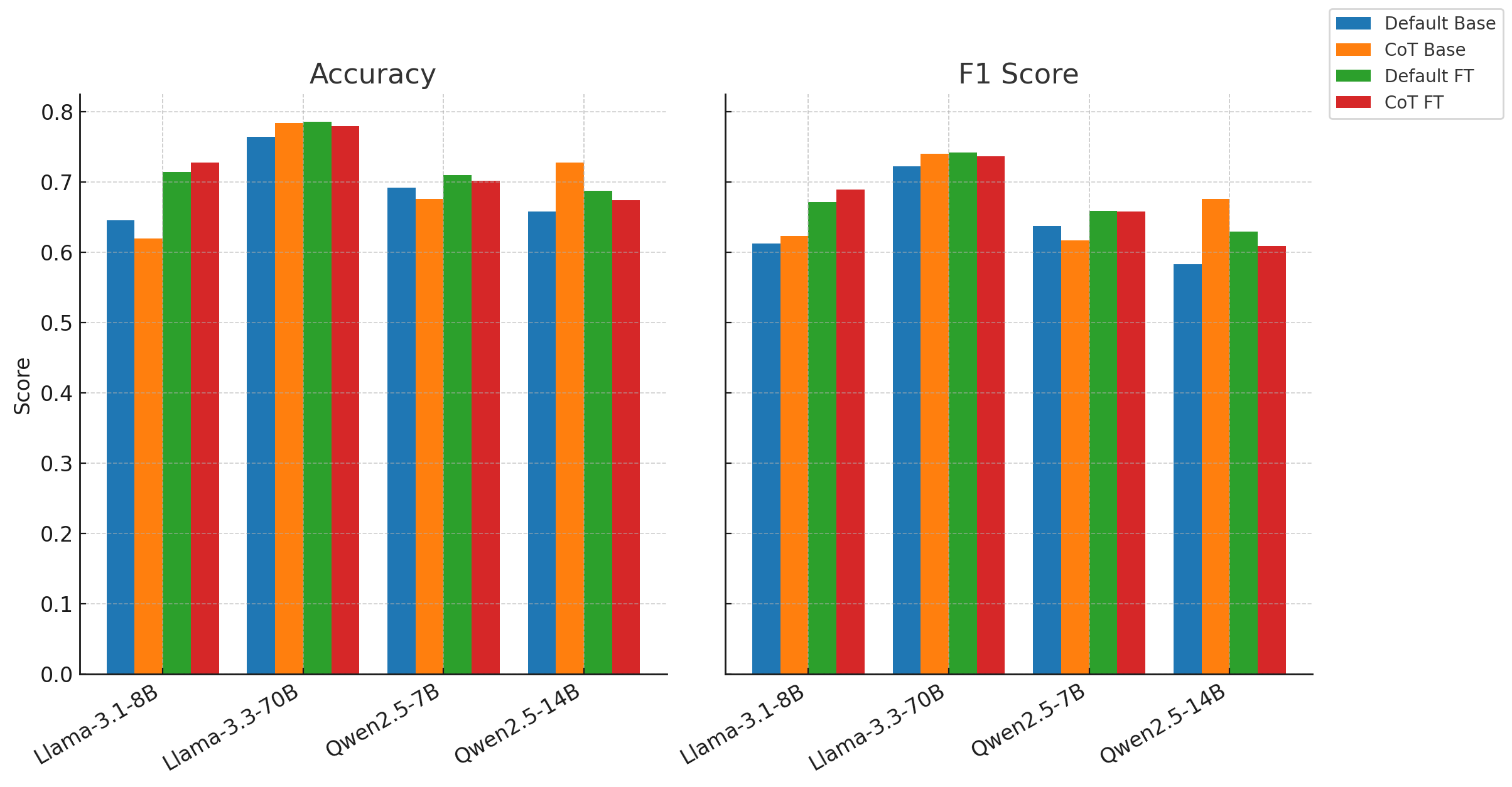}
\caption{Model performance (Accuracy and F1) across different settings: Default vs. CoT prompt; Base vs. Fine-tuned.}
\label{fig:results_barplot}
\end{figure*}

\begin{table}[ht]
\centering
\begin{tabular}{l|cc|cc}
\toprule
\textbf{Model} & \multicolumn{2}{c|}{\textbf{Base}} & \multicolumn{2}{c}{\textbf{Finetune}} \\
 & \textbf{Acc} & \textbf{F1} & \textbf{Acc} & \textbf{F1} \\
\midrule
Llama-3.1-8B (Default) & 0.6460 & 0.6130 & 0.7140 & 0.6716 \\
Llama-3.1-8B (CoT)     & 0.6200 & 0.6236 & 0.7280 & 0.6891 \\
\midrule
Llama-3.3-70B (Default) & 0.7640 & 0.7224 & 0.7860 & 0.7420 \\
Llama-3.3-70B (CoT)     & 0.7840 & 0.7399 & 0.7800 & 0.7366 \\
\midrule
Qwen2.5-7B (Default) & 0.6920 & 0.6372 & 0.7100 & 0.6589 \\
Qwen2.5-7B (CoT)     & 0.6760 & 0.6173 & 0.7020 & 0.6577 \\
\midrule
Qwen2.5-14B (Default) & 0.6580 & 0.5834 & 0.6880 & 0.6298 \\
Qwen2.5-14B (CoT)     & 0.7280 & 0.6760 & 0.6740 & 0.6087 \\
\bottomrule
\end{tabular}
\caption{Accuracy and F1 scores for each model under Default and CoT prompting, before and after instruction fine-tuning.}
\label{tab:all_models_4cols}
\end{table}

\subsection{Comparative Analysis of Model Performance}

Table~\ref{tab:all_models_4cols} and Figure~\ref{fig:results_barplot} summarize the performance of all models under four settings: Base vs. Fine-tuned and Default vs. Chain-of-Thought (CoT) prompting. Three consistent patterns emerge across model families:

\begin{enumerate}
\item \textbf{Instruction fine-tuning leads to consistent improvements}. All models show performance gains after SFT, with accuracy increases ranging from +1.0\% (Qwen2.5-7B CoT) to +8.0\% (Llama-3.1-8B Default). This trend is especially notable under default prompting, where fine-tuning provides clearer gains than CoT.
\item \textbf{CoT prompting improves F1 in base models}. In 3 out of 4 models (Llama-3.1-8B, Llama-3.3-70B, Qwen2.5-14B), adding CoT to base models increases F1, suggesting that CoT can help models handle ambiguous or nuanced biomedical questions. For instance, Llama-3.3-70B shows an F1 increase from 0.7224 (Default) to 0.7399 (CoT) in the base setting.
\item \textbf{CoT + SFT is not always beneficial}. While CoT SFT gives the highest F1 in some settings (e.g., 0.6891 for Llama-3.1-8B), it can hurt performance in others. Notably, Qwen2.5-14B shows a drop in F1 from 0.6760 (Base CoT) to 0.6087 (CoT + SFT), indicating potential misalignment between CoT-style reasoning and instruction-tuned representations for larger models.
\end{enumerate}

\subsection{Model-Specific Insights}

\paragraph{Llama-3.1-8B.}
This mid-sized model exhibits strong gains from both CoT and instruction tuning. Notably, CoT + SFT reaches the highest F1 score among all configurations (0.6891), representing a +7.6\% improvement over the base default (0.6130). This suggests that Llama-3.1-8B benefits from explicit reasoning signals when combined with domain adaptation.

\paragraph{Llama-3.3-70B.}
The largest model in our study already performs well under default prompting (0.7224 F1) and shows a modest improvement with CoT (0.7399). However, post-finetuning, the gains become negligible or slightly negative: CoT + SFT yields 0.7366, slightly lower than the default SFT at 0.7420. This indicates that the model may already internalize reasoning processes, making additional CoT-style supervision unnecessary or even restrictive.

\paragraph{Qwen2.5-7B.}
This model shows stable behavior under both prompting styles, but CoT prompts do not outperform default prompts, neither in base nor in fine-tuned settings. The highest F1 (0.6589) is achieved under default + SFT, while CoT + SFT lags slightly behind (0.6577), suggesting that the added reasoning structure provides marginal utility here.

\paragraph{Qwen2.5-14B.}
Interestingly, this model benefits the most from CoT prompting in the base setting (F1 from 0.5834 to 0.6760, a +9.3\% jump), but fails to maintain this advantage after instruction tuning (F1 drops to 0.6087). This sharp decline highlights a potential misalignment between the model’s internal reasoning and externally imposed CoT formats during SFT.

\subsection{CoT: Help or Hurdle?}

Our findings suggest that Chain-of-Thought prompting is most helpful in zero-shot or base model scenarios, which scaffolds latent reasoning without altering model weights. However, once models are fine-tuned—particularly under instruction-aligned datasets—the benefit of CoT becomes less predictable. For smaller models like Llama-3.1-8B, CoT and instruction tuning appear synergistic. For larger models, the interaction can be neutral or even detrimental.

These results highlight the importance of aligning reasoning style (e.g., CoT) with model scale and tuning objectives. Over-constraining a model with verbose or unnatural reasoning patterns during fine-tuning may hinder its generalization ability.

\section{Conclusion and Future Work}
In this study, we explored the impact of prompt design and instruction fine-tuning on the performance of large language models for biomedical question answering. We find that Chain-of-Thought prompting improves zero-shot performance, while instruction fine-tuning further enhances model quality under default prompts. However, the benefit of CoT fine-tuning is not universal and can occasionally degrade performance, especially in large-scale models.

In future work, we aim to explore:
\begin{itemize}
    \item Multi-stage training with CoT pretraining before full instruction tuning;
    \item Faithfulness and reasoning quality evaluation for Think: outputs;
    \item Expansion to real-world clinical tasks with explainability constraints;
    \item Combining retrieval-based methods (RAG) with CoT prompting for grounded biomedical QA.
\end{itemize}

We hope this work serves as a foundation for robust and explainable medical LLM systems in high-stakes settings.

\bibliographystyle{IEEEtran}
\bibliography{main}

\end{document}